\documentclass[letterpaper, 10 pt, journal, twoside]{IEEEtran}
\usepackage{cite}

\usepackage{bbm}
\usepackage{algorithm}
\usepackage{algorithmicx}
\usepackage{algpseudocode}
\usepackage{amsmath}
\usepackage{mathrsfs}
\usepackage{multirow}
\usepackage{booktabs}
\usepackage{makecell}
\usepackage{array}

\usepackage{amssymb}
\usepackage{mathrsfs}
\usepackage{amsfonts,bm}
\usepackage{algorithm}
\usepackage{color}
\usepackage{cite}
\usepackage{graphics} % for pdf, bitmapped graphics files
\usepackage{epsfig} % for postscript graphics files
\usepackage{times} % assumes new font selection scheme installed
\usepackage{setspace}
\usepackage{algpseudocode}
\usepackage{graphicx}  %插入图片的宏包
\usepackage{float}  %设置图片浮动位置的宏包
\usepackage{subfigure}  %插入多图时用子图显示的宏包
\usepackage{multirow}
\usepackage{rotating}
\usepackage{makecell}
\usepackage{adjustbox}
\usepackage{rotating}
\usepackage[utf8]{inputenc} % allow utf-8 input
\usepackage[T1]{fontenc}    % use 8-bit T1 fonts
\usepackage{hyperref}       % hyperlinks
\usepackage{url}            % simple URL typesetting
\usepackage{booktabs}      % blackboard math symbols
\usepackage{nicefrac}       % compact symbols for 1/2, etc.
\usepackage{microtype}      % microtypography
\usepackage{lipsum}
\usepackage{soul, color, xcolor}
\usepackage{indentfirst}
\usepackage{CJK}
\usepackage{tabularx}

\begin{document}
\title{Goal-Conditioned Reinforcement Learning with Disentanglement-based Reachability Planning}
% \title{REPlan: Disentanglement-based Reachability \\ Planning for Goal-Conditioned RL}
% \title{Goal-Conditioned RL with Reachability Planning and Disentangled Representation}
% \title{REPlan: Reachability Enhanced Planning on Disentangled Representation for Goal-Conditioned Reinforcement Learning}

\author{Zhifeng Qian, Mingyu You$^*$, Hongjun Zhou, Xuanhui Xu and Bin He%
	\thanks{Manuscript received February, 23, 2023; Revised April, 15, 2023; Accepted June, 13, 2023.}
	\thanks{This paper was recommended for publication by Editor Jens Kober upon evaluation of the Associate Editor and Reviewers' comments.}
	\thanks{This work was supported in part by the National Natural Science Foundation of China under Grant 62073244, Shanghai Innovation Action Plan under Grant 20511100500 and Innovation Program of Shanghai Municipal Education Commission (202101070007E00098).}
	\thanks{Z. Qian, M. You, H. Zhou, X. Xu and B. He are with the College of Electronic and Information Engineering, Frontiers Science Center for Intelligent Autonomous Systems, Tongji University, Shanghai 201800, China. }%
	\thanks{*Corresponding author: Mingyu You. (e-mail: myyou@tongji.edu.cn)}
	\thanks{Digital Object Identifier (DOI): see top of this page.}
}

% \author{IEEE Publication Technology,~\IEEEmembership{Staff,~IEEE,}}
        % <-this % stops a space

\markboth{IEEE Robotics and Automation Letters. Preprint Version. Accepted June, 2023}
{Qian \MakeLowercase{\textit{et al.}}: Goal-Conditioned Reinforcement Learning with Disentanglement-based Reachability Planning} 


\maketitle

\begin{abstract}
Goal-Conditioned Reinforcement Learning (GCRL) can enable agents to spontaneously set diverse goals to learn a set of skills. Despite the excellent works proposed in various fields, reaching distant goals in temporally extended tasks remains a challenge for GCRL. Current works tackled this problem by leveraging planning algorithms to plan intermediate subgoals to augment GCRL. Their methods need two crucial requirements: (i) a state representation space to search valid subgoals, and (ii) a distance function to measure the reachability of subgoals. However, they struggle to scale to high-dimensional state space due to their non-compact representations. Moreover, they cannot collect high-quality training data through standard GC policies, which results in an inaccurate distance function. Both affect the efficiency and performance of planning and policy learning. In the paper, we propose a goal-conditioned RL algorithm combined with Disentanglement-based Reachability Planning (REPlan) to solve temporally extended tasks. In REPlan, a Disentangled Representation Module (DRM) is proposed to learn compact representations which disentangle robot poses and object positions from high-dimensional observations in a self-supervised manner. A simple REachability discrimination Module (REM) is also designed to determine the temporal distance of subgoals. Moreover, REM computes intrinsic bonuses to encourage the collection of novel states for training. We evaluate our REPlan in three vision-based simulation tasks and one real-world task. The experiments demonstrate that our REPlan significantly outperforms the prior state-of-the-art methods in solving temporally extended tasks. 
\end{abstract}


%%%%%%%%%%%%%%%%%%%%%%%%%%%%%%%%%%%%%%%%%%%%%%%%%%%%%%%%%%%%%%%%%%%%%%%%%
\section{Introduction}

\IEEEPARstart{I}{ntelligent} agents are expected to master general-purpose skills in open environments.
% \IEEEPARstart{G}
Therefore, Goal-Conditioned Reinforcement Learning (GCRL) is proposed to enable agents to spontaneously set diverse goals to learn a wide range of skills. 
The Goal-Conditioned (GC) policies can generalize skills to different situations and tasks. 
Many excellent works on GCRL have been proposed in various fields, such as robot navigation \cite{shah2021ving, nasiriany2019planning} and manipulation \cite{qian2022weakly, goyal2022ifor}. 

However, GCRL often struggles to reach long-term goals in temporally extended tasks\cite{ijcai2022p770},  since the space to explore becomes large over the horizon, which makes agents difficult to explore high-quality states. In complex environments with high-dimensional observations, e.g. images, such problems are aggravated.
To address this challenge, recent works \cite{chane2021goal, zhang2021c} leverage planning algorithms to enhance GCRL in temporally extended tasks. These methods plan a series of subgoals from the state space, and optimize them based on the distance measure of subgoals, which can serve as a curriculum to improve the learning efficiency of GCRL. 

One crucial requirement of such methods \cite{eysenbach2019search} is to learn a semantically valid and compact representation space, where suitable subgoals can be efficiently searched. The directly planned subgoals in the high-dimensional space may be random noises or invalid images, e.g. unfeasible robot poses. 
To ensure valid subgoals, some works \cite{nasiriany2019planning, chane2021goal} learn a representation space by a variational autoencoder (VAE) \cite{Kingma2014AutoEncodingVB}. However, such representations are non-compact, and inconsistent with the physical properties of entities, as is shown in Section \ref{consist-exp}. Such semantic inconsistency may reduce the efficiency of planning and GC policy learning.

\begin{figure}[t]%htb
	\centering
	\includegraphics[width=7cm]{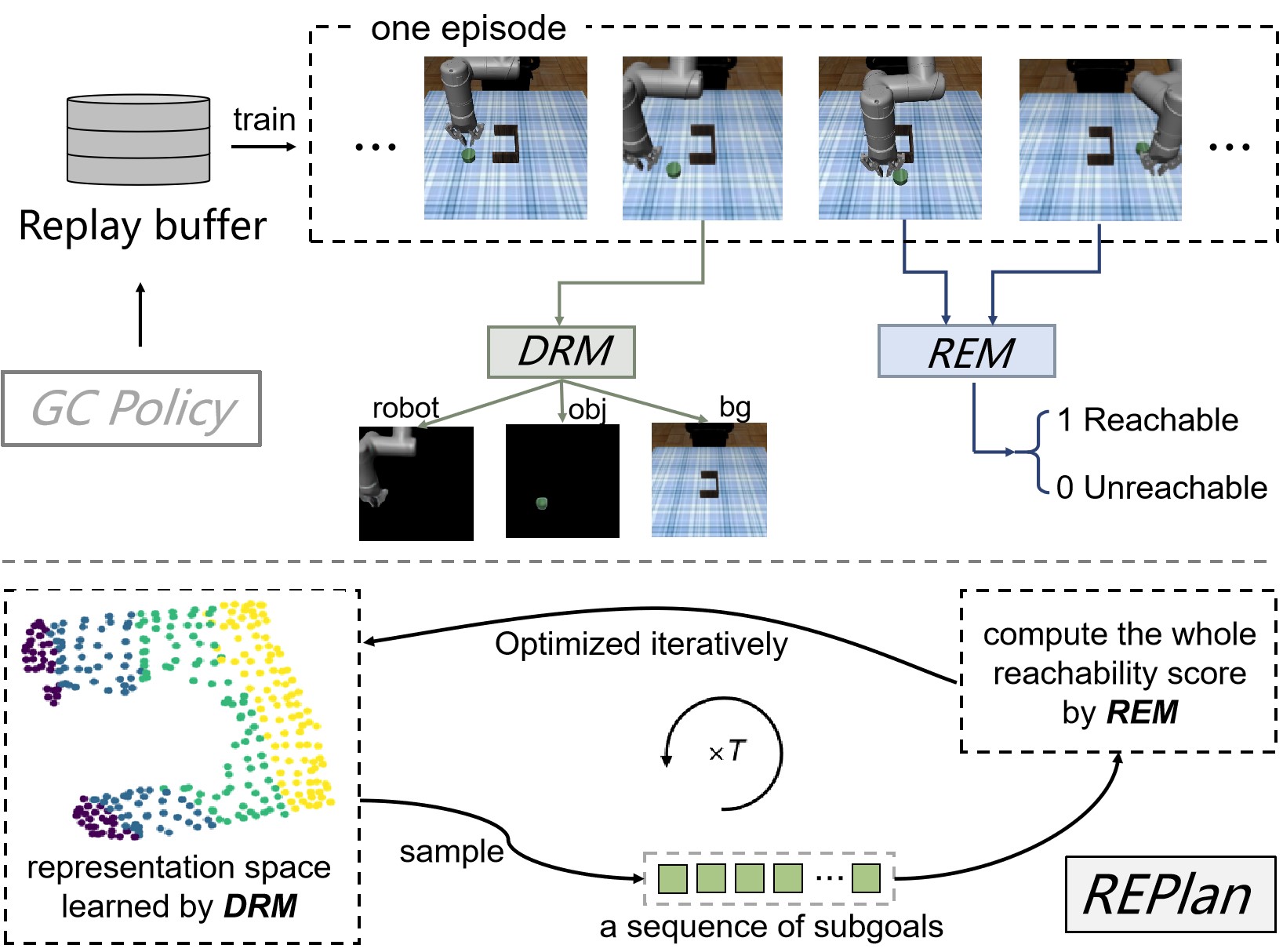}
        \caption{Illustration of our proposed REPlan. In REPlan, DRM aims to learn compact representations to disentangle the robot poses and object positions, while REM is designed to measure the reachability of subgoals. To plan appropriate subgoals efficiently, valid subgoals are sampled from the representation space learned by DRM. Then REPlan computes the whole reachability score by REM. Finally, REPlan optimizes the objective iteratively and outputs an appropriate sequence of subgoals.
        }
	\label{intro}
	 \vspace{-1.6em}
\end{figure}

Another key requirement is to build an accurate distance measure of subgoals, which is used to determine whether a subgoal is reachable from another subgoal. The planner can further compute and optimize the distance of the entire subgoal sequence based on the distance measure. 
% Hafner et al. \cite{hafner2019learning} use the reward prediction model as the distance metric. However, it is laborious to train the prediction model based on complex images, and predicting the entire trajectory may accumulate errors over time, leading to inaccurate estimation. 
Recently, the value function of the GC policy is used as the distance measure in \cite{nasiriany2019planning, eysenbach2019search}. 
However, these methods only construct plans during testing, which means they cannot use subgoals to guide the GC policy to explore high-quality samples during training. 
While the accuracy of the value function depends on the quality of the training samples \cite{zhang2021c}, such methods may result in the estimation deviation of the value function, which is not beneficial to planning and GC policy learning.

To this end, we propose a goal-conditioned RL algorithm combined with Disentanglement-based \textbf{RE}achability \textbf{Plan}ning (\textbf{REPlan}) to solve temporally extended tasks. The illustration of REPlan is shown in Fig. \ref{intro}. 
REPlan is designed to learn a compact representation space where subgoals can be sampled efficiently, and to learn an accurate distance measure that can be used to measure the total quality of subgoals.
A self-supervised \textbf{D}isentangled \textbf{R}epresentation and probabilistic generation \textbf{M}odule (\textbf{DRM}) is proposed to learn a compact representation space, which can disentangle the robot, objects and the background without corresponding mask labels.
Through the proposed inter-branch and intra-branch entropy loss, the manifold structure of the learned representations can be consistent with the robot pose and object position, which greatly improves the planning performance.
In addition, a \textbf{RE}eachability discrimination \textbf{M}odule (\textbf{REM}) is designed to measure the reachability between subgoals. After sampling a sequence of subgoals from the learned disentangled representation space, REPlan iteratively minimizes the whole reachability score computed by REM. 
% REM regards temporally close sample pairs in one episode as positive samples. 
REM converges quickly and can generalize to the unseen distribution. Moreover, REM can compute intrinsic bonuses to encourage exploration and improve the quality of training data, which in turn makes REM more accurate.
We show the data efficiency and advanced performance of REPlan in three simulation tasks and one real-world task, which need temporally extended reasoning.

Our main contributions are as follows:
% to propose a GCRL algorithm combined with planning for solving temporally extended tasks, specifically as follows:
\begin{itemize}
\item We propose REPlan to plan subgoals efficiently to augment GCRL to reach distant goals in temporally extended tasks.
% decompose temporally extended tasks into a series of appropriate subgoals, which is easier to reach for GCRL. 
\item We design DRM to learn representations which can disentangle robot poses and object positions in a self-supervised manner. Such compact representations can improve the efficiency of planning and GC policy learning.
\item We propose REM to measure the reachability of subgoals, which is easy to train and can generalize to the unseen distribution. In addition, REM provides intrinsic bonuses to encourage exploration.
\item We empirically demonstrate that our REPlan outperforms other state-of-the-art methods in three vision-based simulation tasks and one real-world task.
\end{itemize}

%%%%%%%%%%%%%%%%%%%%%%%%%%%%%%%%%%%%%%%%%%%%%%%%%%%%%%%%%%%%%%%%%%%%%%%%%
\section{Related works}

\subsection{Goal-conditioned RL for Temporally Extended Tasks}
% Goal-conditioned reinforcement learning has been studied by many works \cite{kaelbling1993learning, 1999Multi, Schaul2015UniversalVF, colas2022autotelic}. 
Many works on GCRL \cite{kaelbling1993learning, colas2022autotelic} aim to learn GC policies which enable agents to achieve various goals. 
% Schaul et al. \cite{Schaul2015UniversalVF} construct a general value function approximator by decomposing the observation into independent embeddings of state and goal. 
Hindsight Experience Replay (HER) \cite{Andrychowicz2017HindsightER, packer2021hindsight} is often employed to improve the sample efficiency and robustness by relabeling the transition tuples in the replay buffer. While such methods can learn greedy policies to reach nearby goals, they often struggle to solve temporally extended tasks.

To solve this problem, some prior methods \cite{eysenbach2019search, zhang2021world} combine subgoal planning with GCRL. 
In particular, the value function of GCRL is often used to measure the dynamic distances between states for planning. 
SoRB \cite{eysenbach2019search} searches a sequence of subgoals in the replay buffer to guide GCRL to lead to the final goals. 
LEAP \cite{nasiriany2019planning} employs Temporal Difference Models (TDMs) \cite{pong2018temporal}, a goal-conditioned value function, to plan subgoals in the latent space. 
However, the accurate estimation of the value function depends on the quality of the interactive data. These methods only plan subgoals during testing, and can't improve the underlying GC policy to sample higher-quality training data.
To this end, RIS \cite{chane2021goal} performs subgoal search by the high-level policy during training, and uses the Kullback-Leibler (KL) regularization to improve the exploration efficiency. However, the value function is easy to overestimate and unstable during training, which may affect the algorithm performance. C-planning \cite{zhang2021c} employs variational inference to perform a subgoal curriculum to improve the quality of the training data. However, the method excels at dealing with state-based tasks, without additional representation methods designed for vision-based tasks.

\subsection{Representation Learning for Vision-based Control}
Many previous works \cite{qian2022weakly, Srinivas2020CURLCU} have proved that learning an expressive representation can improve the sample efficiency of planning algorithms and RL.  
Srinivas et al. \cite{Srinivas2020CURLCU} combine contrastive learning with model-free RL. 
% Zhang et al. \cite{zhang2020learning} learns task-irrelevant representations using bisimulation metric. However, the physical meanings of these representations are ambiguous. 
Many works \cite{nair2018visual, 9561692} on vision-based control use VAE \cite{Kingma2014AutoEncodingVB} to learn the latent codes of images. However, the learned representations contain redundant information, such as background, and cannot be consistent with the physical attributes of the entity. Later, inspired by compositional scene representation learning in computer vision, recent works learn object-oriented representations in an unsupervised \cite{yuan2022sornet, wang2021roll} or weakly supervised \cite{qian2022weakly} manner for downstream control. However, they only decompose the objects from the background and fail to decompose the complex robots in the manipulation scenes. In contrast, our DRM can not only disentangle the position, texture and mask of the objects but also disentangle the texture and mask of the robot in a self-supervised manner. Our experiments show that our disentangled representations can improve the efficiency and performance of planning.

\subsection{Curiosity-Motivated Exploration}
Curiosity-motivated exploration for RL has been studied in the literature. Task-irrelevant intrinsic rewards are computed based on curiosity to encourage agents to visit a wide range of states. Prediction-based methods \cite{pathak2017curiosity, burda2018exploration} use the prediction error as the intrinsic motivation. Count-based methods \cite{tang2017exploration, joleco} count visited states and reward infrequently visited states. However, the above methods usually don't excel at high-dimensional observation spaces with long episodic time. We take inspiration from the memory-based method \cite{Savinov2019_EC}, which computes a curiosity bonus by training a network to compare the current observation with those in memory. The difference is that our REM is not only used to encourage exploration but also can measure the subgoal quality for planning.

%%%%%%%%%%%%%%%%%%%%%%%%%%%%%%%%%%%%%%%%%%%%%%%%%%%%%%%%%%%%%%%%%%%%%%%%%
\section{Methods}
We propose a goal-conditioned RL algorithm augmented by Disentanglement-based Reachability Planning (REPlan) to solve temporally extended tasks with high-dimensional observations. 
In REPlan, a self-supervised Disentangled Representation and probabilistic generation Module (DRM) is proposed to learn a compact representation space which can disentangle the physical attributes of robots and objects for efficient planning. 
After sampling a sequence of subgoals from the representation space, REPlan computes and optimizes iteratively the reachability score of these subgoals, which is measured by a REachability discrimination Module(REM). To encourage GC policies to collect novel training data, REM can also provide an intrinsic bonus.

In the following, we start by introducing the self-supervised learning of DRM in Section \ref{3.1}. Then Section \ref{3.2} presents the architecture and role of REM. The algorithm summary and details of REPlan are shown in Section \ref{3.3}.

% \begin{figure}[t]
% 	\centering
% 	\includegraphics[width=8.8cm]{img/REPlan.jpg}
%         \caption{Illustration of REPlan exploring, training and testing. During exploring and testing, DRM maps the given initial state and the goal to the corresponding disentangled representations. Then, the planner iteratively optimizes T times to output a series of subgoals based on REM. The GC policy interacts with the environment conditioned on the subgoals, and stores the obtained samples in the replay buffer. During training, DRM, REM and the GC policy are trained on the samples in the replay buffer. In addition, REM encourages exploration by computing an intrinsic bonus to augment each reward. }
% 	\label{REPlan-total}
% 	% \vspace{-1.6em}
% \end{figure}

\subsection{Self-supervised Disentangled Representation for Visual Planning}\label{3.1}
The goal of DRM is to learn a compact representation space, which can ensure that the sampled subgoals are valid. To efficiently plan, the representation should not contain task-irrelevant information. Inspired by previous works \cite{lin2019space, jiang2019scalor} in computer vision, our previous work \cite{qian2022weakly} decomposes object representations from the background for robot control. However, in solving temporally extended tasks, object representations cannot model the process of robot movement and object immobility. 
To this end, DRM can not only disentangle the position, texture and mask of the objects but also disentangle the texture and mask of the robot. Furthermore, our DRM can learn disentangled representations in a self-supervised manner, i.e. without any explicit mask labels.

\begin{figure}[htb]
	\centering
	\includegraphics[width=9cm]{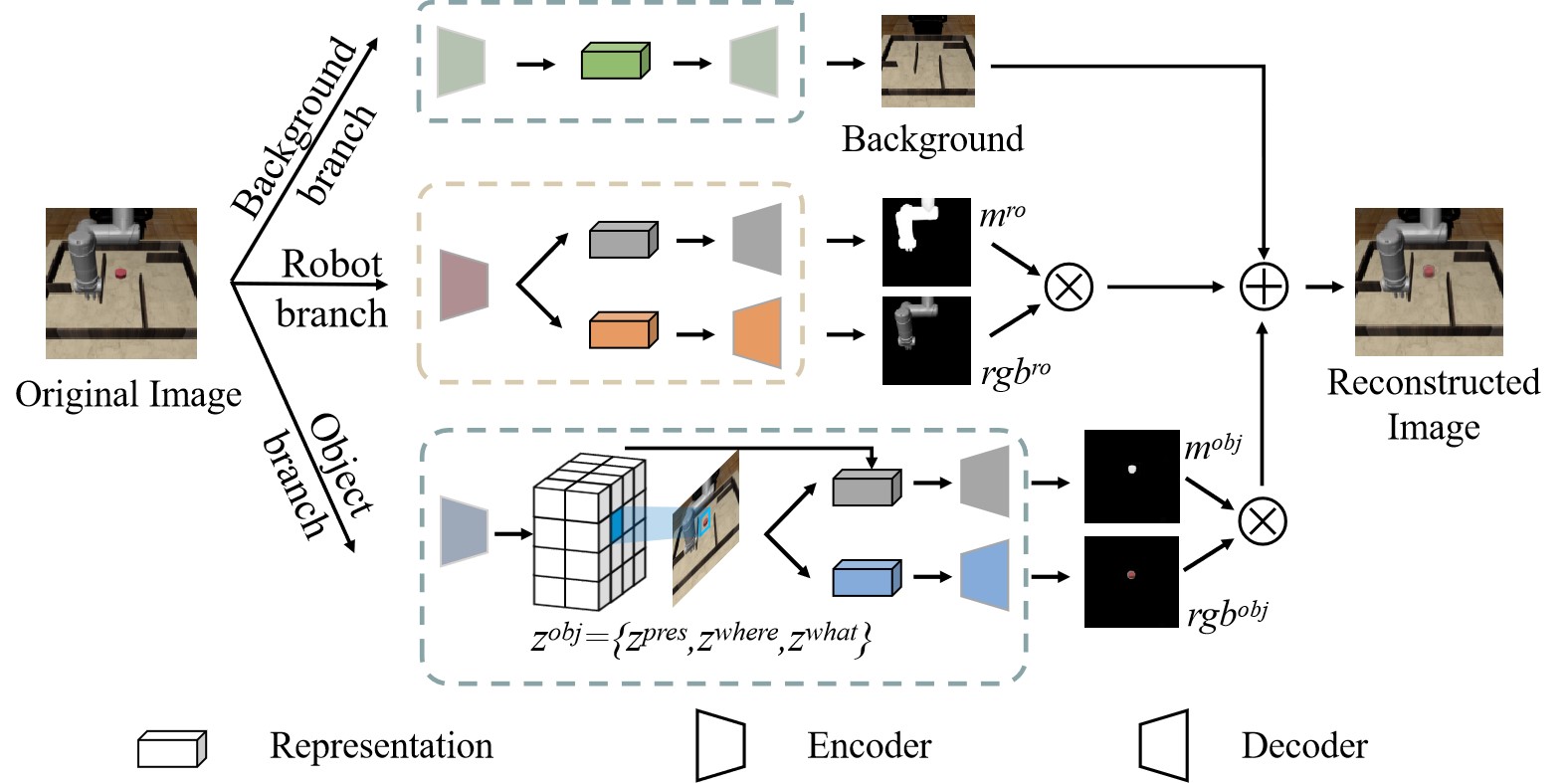}
        \caption{Schematic of DRM. Taking the original image as input, DRM extracts the independent representations of the background, robot and objects through three branches. The representation of the robot or objects can be decoded into masks and textures. Finally, DRM uses a pixel-wise mixture model to reconstruct the complete image.}
	\label{DRM}
	% \vspace{-1.6em}
\end{figure}

The schematic of DRM is shown in Fig. \ref{DRM}. DRM decomposes the manipulation scene into three independent representations: robot $z^{ro}$, object $z^{obj}$ and background $z^{bg}$. On top of them, DRM uses a pixel-wise mixture model to reconstruct the complete image:
\begin{equation}
\label{3.2.1}
\begin{aligned}
  p\left(s \mid z^{ro},z^{obj}, z^{bg}\right) = 
  \underbrace{m^{ro}\left(z^{ro,m} \right)  p\left(s \mid z^{ro,rgb}\right)}_{\text {Robot}}\\
+\underbrace{m^{obj}\left(z^{obj,m} \right)  p\left(s \mid z^{obj,rgb}\right)}_{\text {Foreground }} 
+\underbrace{m^{bg} p\left(s \mid z^{bg}\right)}_{\text {Background }}          
\end{aligned}
\end{equation}
where the mixing probability $m\left(z \right)$ is the mask computed based on $z$, and $m^{bg}=1-m^{ro}\left(z^{ro,m} \right)- m^{obj}\left(z^{obj,m} \right)$. DRM models the RGB distribution $p\left(s \mid z^{ro,rgb}\right)$, $p\left(s \mid z^{obj,rgb}\right)$ and $p\left(s \mid z^{bg,rgb}\right)$ as a Gaussian $\mathcal{N}(\mu^{ro} ,\sigma^{ro})$, $\mathcal{N}(\mu^{obj} ,\sigma^{obj})$and $\mathcal{N} (\mu^{bg} ,\sigma^{bg})$. $\sigma$ is fixed while $\mu$ can be learned.
% as dictated by \cite{lin2019space}.

To model the objects in the scene, DRM divides the observation $s \in \mathbb{R}^{3 \times 128 \times 128}$ into $H \times W$ cells. Each cell models a structured representation $z^{obj,rgb}=\{z^{pres},z^{what},z^{where}\}$, which is similar to \cite{lin2019space}. $z^{pres}$ is a binary value to indicate whether the cell has any object. $z^{what}$ is used to reconstruct the object glimpse and its mask. Then $z^{where}$ is used as the affine transformation parameter of Spatial Transformer \cite{10.5555/2969442.2969465} to transform the glimpse to the original resolution. 
% $z_{depth}$ points out the depth of the object if occlusion happens. 
We model $z^{pres}$ as a Bernoulli distribution while other representations as a Gaussian distribution. So DRM implements $z_{obj}$ as
\begin{equation}
    \label{z-obj}
    p\left(z^{obj,rgb}\right) = \coprod_{i=1}^{H*W}p\left(z_i^{pres}\right)\left(p\left(z_i^{where}\right) p\left(z_i^{what}\right)\right)^{z_i^{pres}}
\end{equation}

We train DRM by optimizing an evidence lower bound(ELBO) \cite{Kingma2014AutoEncodingVB}, which is given by
\begin{equation}
    \label{elbo}
    \begin{aligned}  
\mathcal{L}_{elbo}(x) = &\mathbb{E} _{q(z^{ro},z^{obj},z^{bg}|x)}[logp(x|z^{ro},z^{obj},z^{bg})\\
&- \mathbb{KL}(q(z^{bg}|x)||p(z^{bg}|x))\\
&-\sum_{i=1}^{H*W}\mathbb{KL}(q(z_i^{obj}|x)||p(z_i^{obj}|x)) \\
&- \mathbb{KL}(q(z^{bg}|x)||p(z^{bg}|x))
]
    \end{aligned}
\end{equation}
We reparameterize the continuous variables and model the discrete variables using the Gumbel-Softmax distribution. %\cite{Kingma2014AutoEncodingVB} \cite{jang2017categorical}. 

Different from previous works in computer vision, which only disentangle the object and the background, DRM also disentangles the robot.
Since it is difficult to model robots with huge changes by the cell of the object branch, DRM adds an additional robot branch. 
To prevent confusion between the robot branch and the object branch learning, we introduce an inter-branch entropy loss, which is denoted as
\begin{equation}
    \label{entropy}
    \mathcal{L}_{H_{inter}} = || \sum_{i=1}^{H*W}m^{obj}_i m^{ro} ||_1
\end{equation}
% where $SM$ denotes the softmax operation in pixel dimension. 
To penalize multiple cells for detecting the same object, we introduce an intra-branch entropy loss, which is denoted as
\begin{equation}
    \label{entropy2}
    \mathcal{L}_{H_{intra}} = \sum_{i=1}^{H*W} \sum_k^K-m_{i,k}^{obj}log m_{i,k}^{obj}
\end{equation}
where $K$ denotes the detected objects.
To summarize, the total loss function is the sum of each loss weighted by the relevant hyperparameters, which is denoted as
\begin{equation}
    \label{total-loss}
    \mathcal{L}_{DRM} = \mathcal{L}_{elbo}(x) + \alpha_{inter}\mathcal{L}_{H_{inter}} + \alpha_{intra}\mathcal{L}_{H_{intra}}
\end{equation}

\subsection{Planning with Reachability Discrimination Module}\label{3.2}

\textbf{Planning by reachability.} Appropriate subgoals should be reachable between neighbors. If the GC policy can reach the next subgoal given the current state, the agent can reach the long-term goal by reaching each short-term goal. So how can we measure whether those adjacent subgoals are reachable?

\begin{figure}[t]
	\centering
	\includegraphics[width=6cm]{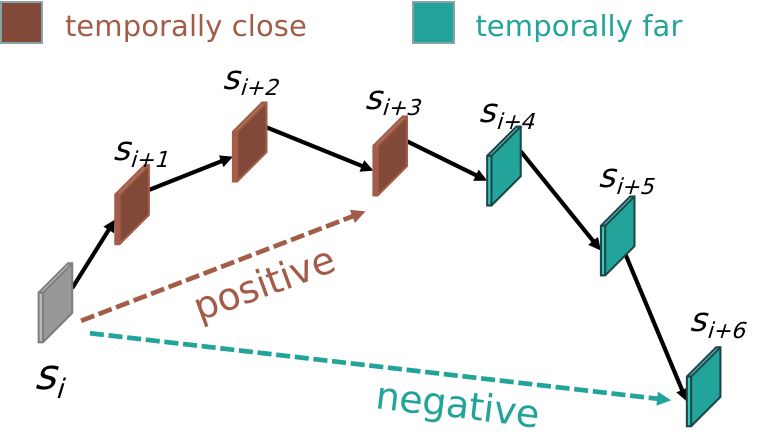}
        \caption{REM is trained based on the interaction trajectories of the GC policy. We regard the observations temporally close to each other in the trajectory as positive pairs, while temporally far observations as negative pairs.}
	\label{RD-train}
	 \vspace{-1.6em}
\end{figure}

Accordingly, REM is proposed to measure the reachability of the adjacent subgoals. 

REM regards predicting whether two states are reachable as a binary classification task, and there are two categories of reachability and unreachability.
Given two observations $s_{i}$ and $s_{j}$, REM predicts the reachable probabilities of the current policy from $s_{i}$ to $s_{j}$.  

More generally, given a high-level trajectory composed of $N$ sequential intermediate subgoals $z_{sg_{1:N}}=z_{sg_1},z_{sg_2},...,z_{sg_N}$, we define the episodic reachability as
\begin{equation}
    \label{loss-RD}
    \overrightarrow{R}(z_{s_0},z_{sg_{1:N}},z_{g}) = \sum_{n=0}^{N}R(z_{sg_n}, z_{sg_{n+1}})
\end{equation}
where $z_{sg_0}$ denotes the initial observation $z_{s_0}$ and $z_{sg_{N+1}}$ denotes the true goal for simplicity of notation. $R$ denotes the REM.
The episodic reachability $\overrightarrow{R}$ provides a measure of a plan's overall reachability. So REPlan is optimized by minimizing the negative norm of the episodic reachability, which is denoted as
\begin{equation}
    \label{loss-plan}
    L(z_{sg_{0:N+1}}) = -||\overrightarrow{R}(z_{sg_{0:N+1}})||_2
\end{equation}

% The architecture of REM is shown in \hl{Fig.} \ref{RD}. 
\textbf{Architecture and Training.} REM is regarded as a classifier consisting of three fully connected layers, which are simple and easy to train. Given two observations $s_i$ and $s_j$, we first use the siamese DRM to extract the disentangled representations $z_i$ and $z_j$. Note that we only use the corresponding robot mask representation $z^{ro, m}$ and the object position representation $z^{obj, where}$, which contain task-relevant information. 
Then REM predicts the reachability probability between two observations. 

The training method is shown in Fig. \ref{RD-train}.
Two states are sampled in a trajectory from the replay buffer and a fixed step distance is used to distinguish whether they are positive or negative samples. In our setting, the step distance is set to 80. In addition, we select a binary cross entropy loss function, which is commonly used in binary classification tasks, to optimize REM:
\begin{equation}
    \label{loss-REM}
    L_{R} = -y \log \hat{y}+(1-y) \log (1-\hat{y})
\end{equation}
where $y$ is the ground truth and $\hat{y}$ is the prediction.

Note that REM is trained on interaction samples of the GC policy. As the capability of the policy is gradually improved, the training distribution of REM continues to change. It requires REM to be simple and easy to train. Therefore, the siamese DRM extracts compact and expressive representations and fixes the gradient during training REM.  

\textbf{Curiosity Module (CM).} 
Another design of REM is to provide an intrinsic bonus to improve the quality of the collected data. 
We draw inspiration from curiosity-based exploration for RL \cite{Savinov2019_EC}. We use a REM-based curiosity module (CM) to reward novel states, which is shown in Fig. \ref{bonus}. 
CM provides positive rewards when the reachability scores of the current state and states in the memory buffer are lower than the novelty threshold. 
% The representations in the memory buffer $MB={z_1,z_2...,z_M}$ are the representations of the states visited by the agent in previous episodes. 

\begin{figure}[t]
	\centering
	\includegraphics[width=7cm]{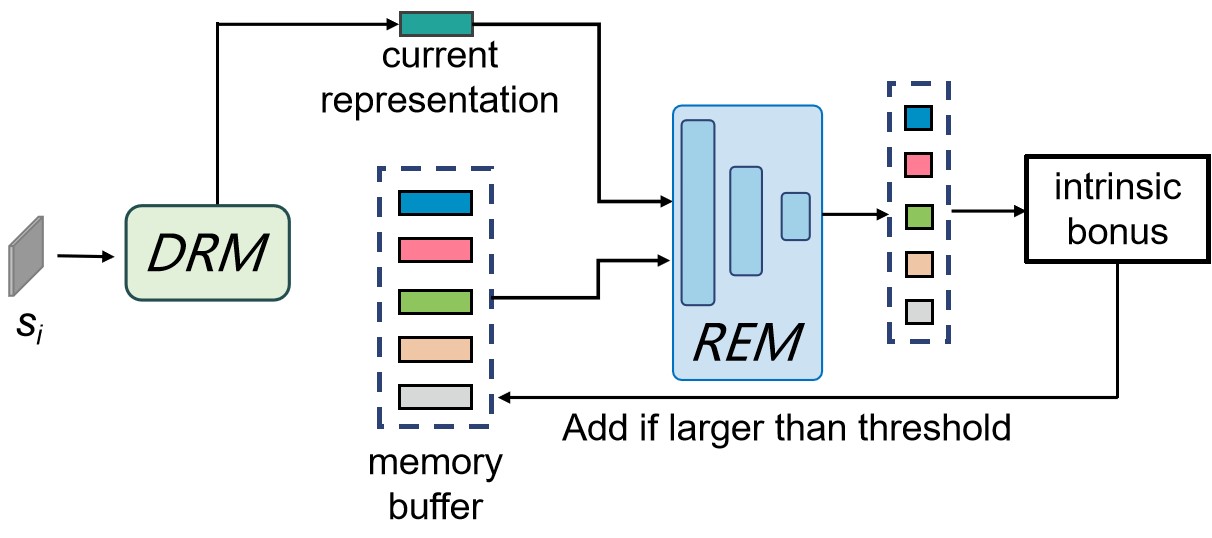}
        \caption{The curiosity module is used to compute the intrinsic bonus to encourage exploration. CM gives positive rewards to novel states by comparing the reachability between the current representation and the representations in the memory buffer.}
	\label{bonus}
	 \vspace{-1.6em}
\end{figure}

\begin{figure*}[t]
	\centering
	\includegraphics[width=16.5cm]{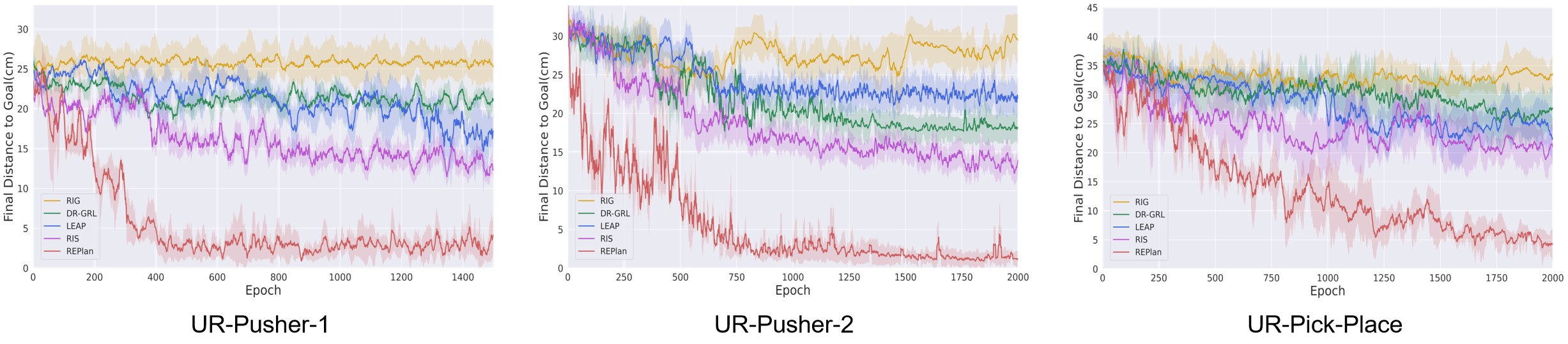}
        \caption{Comparison with several state-of-the-art methods on three vision-based temporally extended tasks \emph{UR-Pusher-1}, \emph{UR-Pusher-2} and \emph{UR-Pick-Place}. The goals of the first two tasks are to push one puck to the goal position with different complex obstacles, while the goal of the last task is to pick up the two cubes and place them on the corresponding cubes.
        % The diagrams of the front view and the agent view are shown. 
        As is shown, our REPlan outperforms other methods in both sample efficiency and performance on three tasks.
        }
	\label{task12}
	\vspace{-1.3em}
\end{figure*}

Given a current observation $S_i$, DRM extracts the corresponding representation $z_i$.
REM compares $z_i$ with those in the memory buffer to obtain reachability scores. 
% To encourage the agent to explore the unseen states, the reachability scores are expected to be minimized. So 
We denote the intrinsic bonus $b^i$ as
\begin{equation}
    \label{bonus}
    b^i = \beta -\alpha MAX[R(z_i, z_m)], m=1:M
\end{equation}
Empirically, we choose $\beta = 0.2$ and $\alpha=0.8$. The memory buffer has a limited capacity $M$. 
After computing the intrinsic bonus, the representation is added to the memory buffer if the bonus is larger than the novelty threshold $b_{novelty}=0$.
When the memory capacity overflows, the new representation replaces a random old one with a random probability.

We use the intrinsic bonus $b_t^i$ to enhance the extrinsic reward $t_t^e$ and improve the efficiency of GCRL. External rewards $t_t^e$ are computed by the representation distance between the current state and the goal. The rewards are computed as
%described in Section \ref{3.3}.
\begin{equation}
    \label{bonus-reward}
    r_t = t_t^e + b_t^i
\end{equation}
\begin{equation}
    \label{e_rew}
    t_t^e = ||z_t - z_g||_1
\end{equation}
\subsection{Summary of Reachability Enhanced Planning with GCRL}\label{3.3}

We summarize the whole algorithm of REPlan in Algorithm. \ref{algorithm}. We first collect data through random exploration and train DRM by Equ. \ref{total-loss}. Then given the initial state $s_0$ and one goal $g$, we use Cross-Entropy Method \cite{de2005tutorial}, a gradient-free optimization algorithm, to optimize Equ. \ref{loss-plan} to choose $N$ subgoals with maximum reachability. Once the subgoals are obtained, we only take the first subgoal as the input of the GC policy. After step $t_n$, we repeat the above procedure: we replan the $N-1$ subgoals and give the first subgoal to the GC policy to interact with the environment $t_n$ times.

\begin{algorithm}[htp]
	\caption{REPlan with GCRL}
	\label{algorithm}
	\begin{algorithmic}[1]
		\State Train DRM with randomly explored samples by Equ. \ref{total-loss}.
            \State Initialize REM $R_{\theta}$, GC policy $\pi_\phi$.
            \State Initialize the max episodes $E$, the number of subgoals $N$.
		\State Sample the initial state $s_0 \sim \rho_0$, and the goal $g \sim \rho_g$.
		\State \textbf{for} $i = 1:E$ episodes \textbf{do}
  
		\State \quad \textbf{for} $n = 1:N$ \textbf{do}
		\State   \quad \quad Optimize Equ. \ref{loss-plan} to choose subgoal representations $z_{sg_n},...,z_{sg_N}$ using REM if $n<N$.
  
		\State \quad \quad \textbf{for} $t=1:t_{N} $ \textbf{do}
		\State  \quad \quad \quad Execute $a_t$ using GC policy $\pi(\cdot|z_{t},z_{sg_n})$.
            \State  \quad \quad \quad Augmenting tuple $[z_t, a_t, z_{t+1}, r_t,]$ by Equ. \ref{bonus-reward}.
            \State  \quad \quad \quad Store the tuple in the replay buffer.
		\State   \quad \quad \textbf{end for}
		\State  \quad \textbf{end for}			
            \State    \quad Sample trajectories to train REM by Equ. \ref{loss-REM}.
		\State 	  \quad Sample tuples to train the GC policy $\pi_\phi$.
		\State \textbf{end for}		
		% \State\textbf{return} {$\theta$}
	\end{algorithmic}
\end{algorithm}

In principle, GCRL can model the variable goal $g$ into any standard reinforcement learning algorithm. We choose the twin delayed deep deterministic policy gradient algorithm (TD3) \cite{fujimoto2018addressing} which is an effective off-policy RL algorithm. 
We also use Hindsight Experience Replay (HER) \cite{Andrychowicz2017HindsightER} to relabel the tuples with the future and generated goals to accelerate policy learning.
In this work, we set the number of subgoals $N$ as a hyperparameter of the task. We compute the maximum horizon between adjacent subgoals $t_n=T_{max}/(N+1)$. 

%%%%%%%%%%%%%%%%%%%%%%%%%%%%%%%%%%%%%%%%%%%%%%%%%%%%%%%%%%%%%%%%%%%%%%%%%%%%%%%%%%%%%%%%%%%%%%%%%%%%%%%%%%%%%%%%%%%%%%%%%%%%%%%%%%%%%%%%%%%%%%%%%%%%%%%%%%%%%%%%%%%%%%%%%%%%%%%%%%%%%%%%%%%%%%%%%%%%%%%%%%%%%%%%%%%%%%%%%%%%%%%%%%%%
\section{Experiments}\label{experiments}
To verify the effectiveness of our REPlan, we perform experiments in three vision-based simulation tasks and one real-world task. All these tasks require agents to achieve the long-term goal with temporally extended reasoning.
We first introduce our experimental setup in Section \ref{setup}. 
In the following sections, our experiments study the following questions:
(1) How does our REPlan compare with prior state-of-the-art works in goal-conditioned reinforcement learning?
(2) Does DRM in REPlan disentangle the components of the observation in a self-supervised manner?
(3) Can the learned disentangled representations maintain
consistency with the physical attributes of entities such as object positions and robot poses?
(4) Can REM in REPlan reflect the reachability of two states?
(5) Is each design of our REPlan really effective for solving vision-based temporally extended tasks? 
(6) Can our REPlan transfer efficiently from simulation to the real world?

\subsection{Experiment Setup}\label{setup}
\textbf{Vision-based Robotic Manipulation Tasks.}
We perform experiments on three complex robotic manipulation tasks in simulation and one real-world task. For the sake of clarity, we denote them as \emph{UR-Pusher-1}, \emph{UR-Pusher-2}, \emph{UR-Pick-Place} and \emph{Real-Pusher}. The tasks are based on Robosuite \cite{zhu2020robosuite} powered by the MuJoCo physics engine \cite{todorov2012mujoco}. The simulation tasks include two tasks of pushing the puck around obstacles and one task of continuously picking up and placing two cubes, while \emph{Real-Pusher} task is similar to \emph{UR-Pusher-1}. Therefore a greedy goal-reaching policy can hardly reach the goal directly from the initial state. It requires the agent to perform temporally extended reasoning.
%The observations obtained from the environment are RGB images, which are reshaped to 128 * 128 resolutions. 
%In particular, the translucent object represents the desired goal, which is unobservable to the agent. 
%During training and evaluation, the initial object state and the goal are randomly sampled. 

\textbf{Baselines.}
We compare our approach to prior state-of-the-art GCRL methods. \textbf{RIG} \cite{nair2018visual}: combine TD3 \cite{fujimoto2018addressing} with HER \cite{Andrychowicz2017HindsightER} for GCRL. RIG also trains a VAE \cite{Kingma2014AutoEncodingVB}, which is used to encode observations, generate unseen goals to relabel samples, and compute the negative Euclidean distance in the representation space as rewards. We regard RIG as a baseline since RIG is a greedy goal-reaching policy without planning subgoals. 
\textbf{DR-GRL} \cite{qian2022weakly}: further learns compact representations capable of disentangling different attributes of objects, which are then utilized for state abstraction and reward function calculation of the GC policies.
\textbf{LEAP} \cite{nasiriany2019planning}: uses VAE \cite{Kingma2014AutoEncodingVB} and Cross-Entropy Method \cite{de2005tutorial} to plan a series of subgoals, which the low-level GC policy need to reach one by one. In addition, LEAP takes the temporal difference models (TDMs) \cite{pong2018temporal}, a time-varying goal-conditioned value function, as the reward. \textbf{RIS} \cite{chane2021goal}: predicts subgoals through the high-level policy, and use the Kullback-Leibler(KL) regularization to help the underlying GC policy learning. For a fair comparison, we select TD3 as the RL algorithm to train REPlan, RIG, LEAP and RIS. All the methods are trained on 4 NVIDIA TITAN X GPUs.

\subsection{Comparison with Prior GCRL Methods}\label{compare}
To answer the first question, we compare our REPlan to several state-of-the-art methods in solving vision-based temporally extended tasks. 
%Since different methods have different definitions of reward functions, 
To clearly and fairly measure the task completion, we use the European distance between the object and the goal on the table at the end of an episode. We train each method for 1500 epochs on task 1 and 2000 epochs on task 2. Each method is trained with 5 seeds. In each epoch, the agent must interact with the environment for 5 episodes, and the max steps in an episode are 500. 
% More implementation details can be found in the Appendix[].

The experiment results are shown in Fig. \ref{task12}. Dark lines represent the mean value and light areas represent the confidence interval.
As is shown, RIG struggles to reach the goal, since RIG is a greedy goal-reaching policy, and uses the Euclidean distance in the VAE feature space as the reward function, which is not consistent with the physical attributes of the robot and the object. RIG cannot perform long-term reasoning, and always pushes the object into obstacle traps. 
DR-GRL is slightly improved compared to RIG, due to the learned object-centric representations, but still hardly reaches distant goals.
LEAP and RIS make progress on three tasks since both of them plan over subgoals to reach distant goals. However, we find that their inefficient exploration in such complex environments leads to inaccurate distance measures, which affects the performance of the algorithm.
Compared to the RIG, DR-GRL, LEAP and RIS, our REPlan converges faster and can reach long-term goals within a short distance. 

\begin{figure}[t]
	\centering
	\includegraphics[width=8cm]{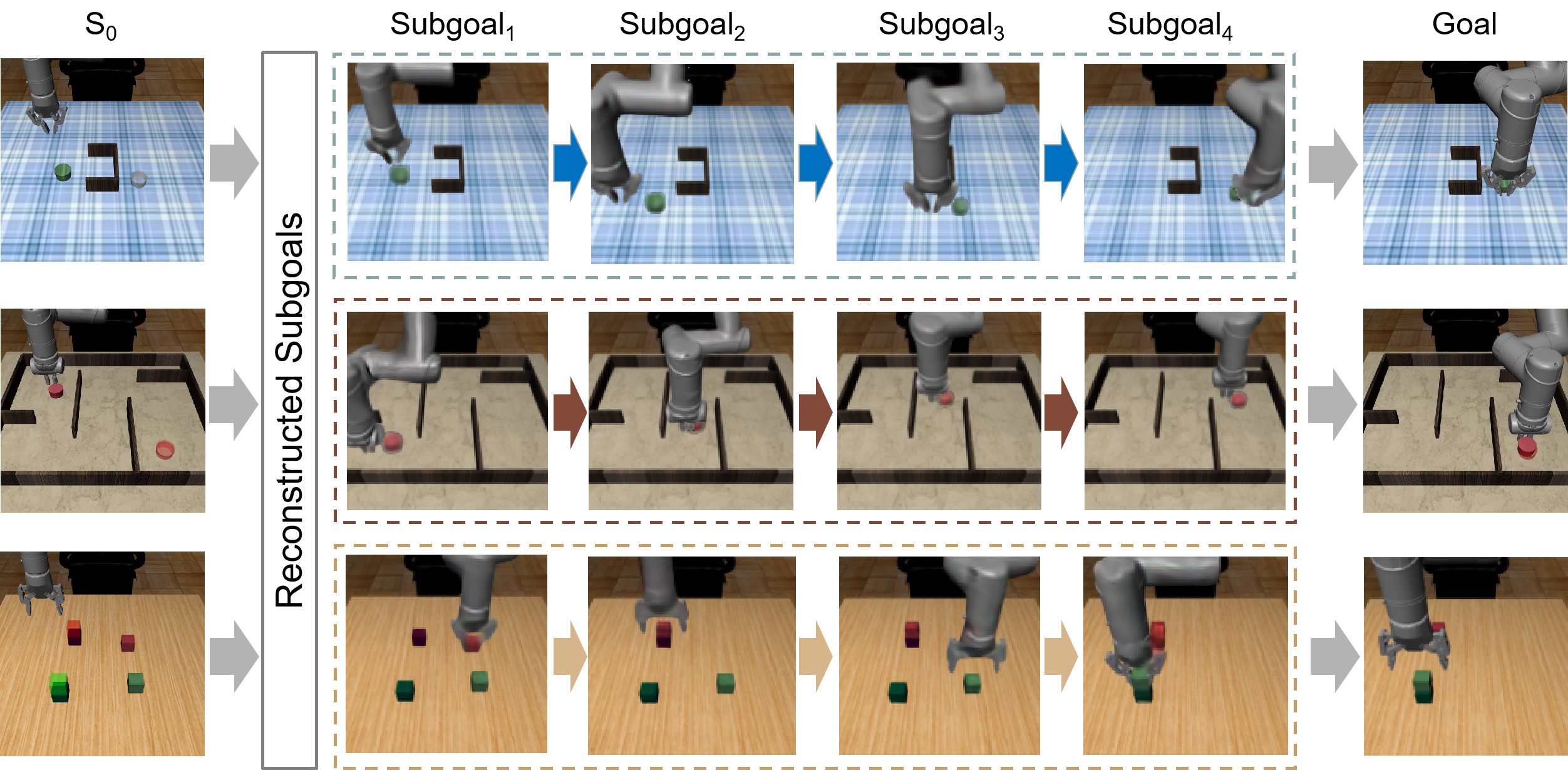}
        \caption{Visualization of the reconstructed subgoals generated by our method. We use DRM to decode the generated goal representations (including robot and object) with the background into the images. }
	\label{subgoal-pic}
	 \vspace{-1.6em}
\end{figure}

We visualize the subgoals generated by our REPlan in Fig. \ref{subgoal-pic}. We use DRM to decode the generated goal representations (including robot and object) with the background. Take task 1 as an example. Given the initial state $S_0$ and the final goal $Goal$, the agent plans a meaningful and reachable subgoal sequence: at $Subgoal_1$, the agent reaches near the object. At $Subgoal_2$, it pushes the object downward. Note that since it is going to push the object to the right next, the agent will move to the left of the object to make it easier to reach the next subgoal. At $Subgoal_3$, it pushes the object to the right below the goal. At $Subgoal_4$, it goes under the object to prepare for the final push to the goal.

\subsection{Disentanglement and Reconstruction of DRM}\label{section-DRM}
To answer the second question, we train DRM on the dataset collected by the random policy in three simulation tasks and one real-world task. The dataset contains different object positions and robot arm poses, as well as different obstacles and different table background textures. In particular, the dataset does not contain any annotation of object coordinates or robot poses. DRM is optimized only by the reconstruction loss and other auxiliary loss functions \ref{total-loss} in a self-supervised manner.

\begin{figure}[t]
	\centering
	\includegraphics[width=7.cm]{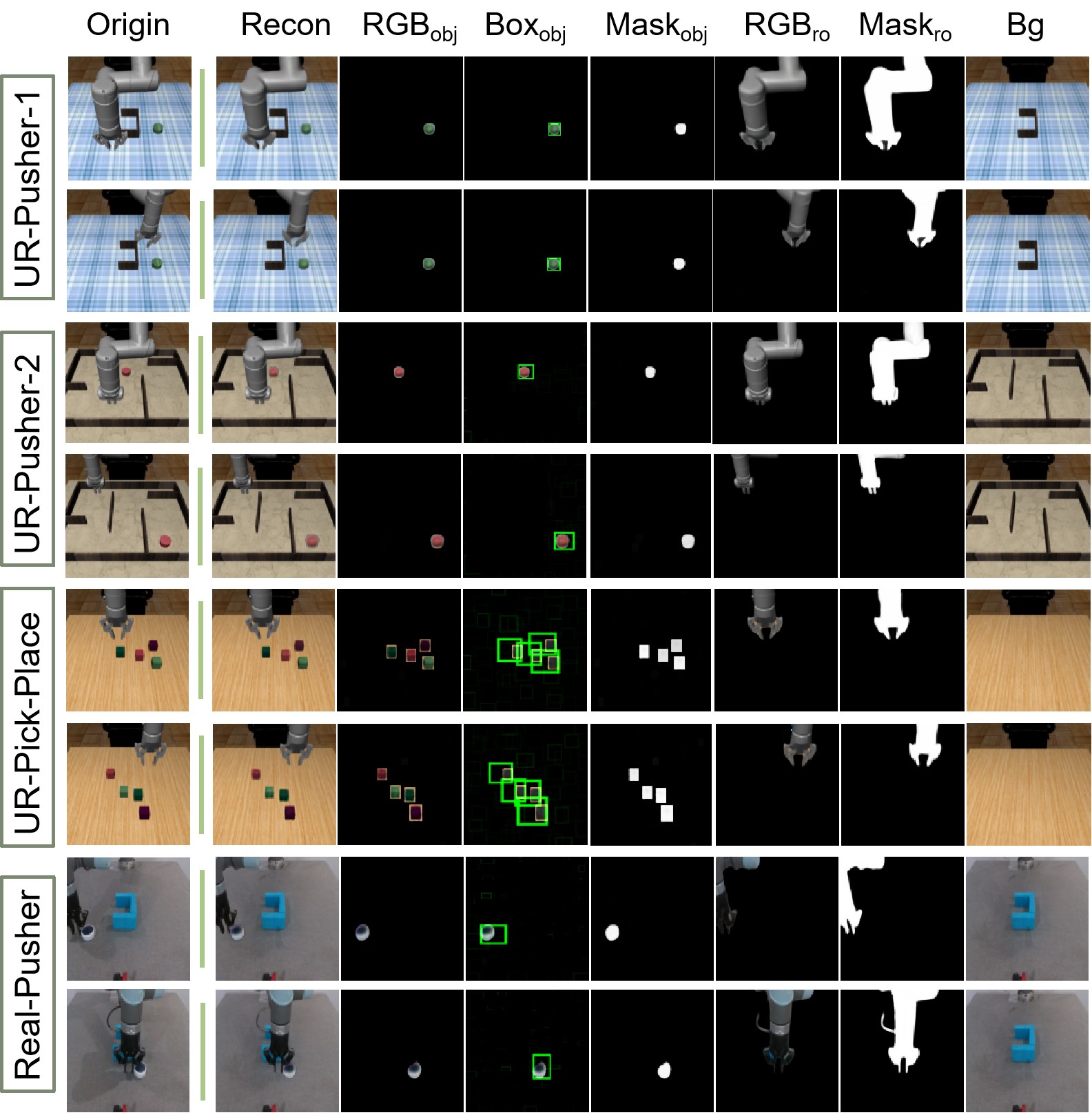}
        \caption{Visualization of the reconstructed observations by DRM. Two examples of each task are shown here. As is shown, DRM can disentangle the robot, the object and the background from images in a self-supervised manner.}
	\label{fenge-pic}
	 \vspace{-1.6em}
\end{figure}

The qualitative results are given in Fig. \ref{fenge-pic}. The first column is the original observations, and the following columns are the overall reconstructed images and the glimpse and the mask of each component. 
As is shown, DRM can clearly disentangle the object and robot from the complex background in both simulation and the real world. Through the spatial transformer module, DRM can accurately predict the object bounding box and generate clear textures and masks in the bounding box. 
% In fact, DRM can also disentangle multiple objects with different shapes and textures in parallel, but this is not the focus of this paper.

\subsection{Consistency of Disentangled Representation and Entity 
Physical Attribute} \label{consist-exp}

\begin{figure*}[t]
	\centering
	\includegraphics[width=16cm]{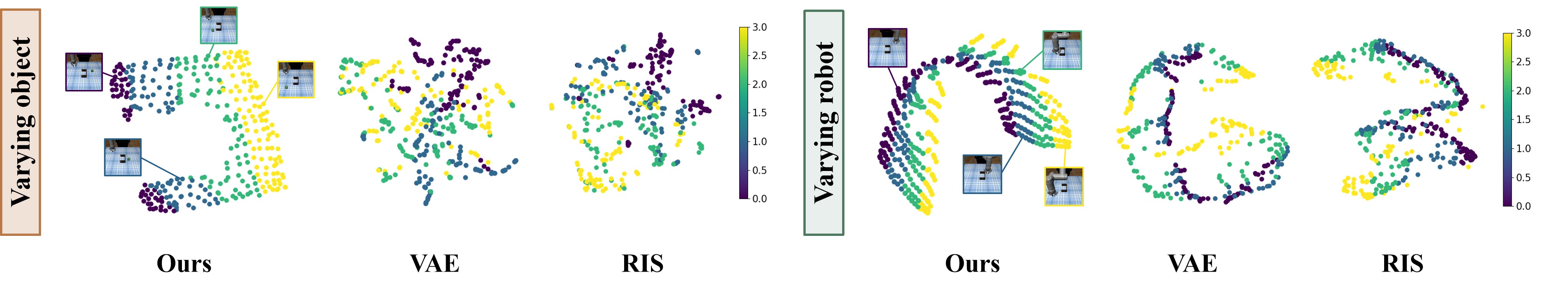}
        \caption{Visualizations of the observation representations learned by different methods after the dimensionality reduction. T-SNE is used to reduce the dimension of these representations to two. We expect to show whether the learned representations are consistent with the object positions and the robot poses in the observations. Therefore, we uniformly sample about 10 * 10 positions of the cups \textbf{(Left)} and about 10 * 10 poses of the robot \textbf{(Right)} to show the different types of observation changes. We use different colors to indicate the changes, where label 0 indicates the upper right of the table and label 3 indicates the lower left.}
	\label{sandian}
	 \vspace{-1.6em}
\end{figure*}

To answer the third question of whether the learned representations are consistent with the object positions and the robot poses, we visualize the representations learned by different methods in Fig. \ref{sandian}. RIG and LEAP use VAE to encode the observation, while RIS optimizes the image encoder and the control policy in an end-to-end manner through reinforcement learning.
In addition, DR-GRL learns object-centric representations similar to those of our REPlan, but DR-GRL cannot learn robot representations.
T-SNE \cite{Maaten2008VisualizingDU} is used to reduce the representation dimension to 2. 

We use observations in \emph{UR-Pusher-1} as examples.
In the left sub-figure, we use points to represent 20*20 different object positions, while we sample 20*20 different robot poses in the right sub-figure. For clarity, we do not sample the object positions in the middle of obstacles. 
We use colors to represent changes in object positions and robot poses. 
It is worth noting that our method can disentangle the object and the robot, while other methods can only encode the overall image. So to reduce the influence of the entity's physical attributes on the representation of other methods, we fix the robot arm at the upper left corner in the left experiment and remove objects in the right experiment. As is shown, the representations of VAE and RIS are relatively messier and overlap. In addition, these representation distances do not change monotonously with the entity attributes. For example, when the distance between the object and the goal gets smaller, the distance between features may become larger. 

In contrast, the disentangled representations learned by DRM are consistent with object positions both horizontally and vertically. As is shown, our DRM can well learn the manifold structure of object positions and robot poses in the representation space.  
% As proved by \cite{qian2022weakly}, such efficient representations can improve the sample efficiency of GCRL.
As further proved by our experiments in section \ref{ablation_each}, such efficient representations can improve the performance of REPlan. In addition, the representations can effectively help reduce the training difficulty of REM.
% are used to encode observations and calculate reward functions, which 
% can improve the sample efficiency of GCRL. 
\vspace{-1.2em}

\subsection{Performance of REM}

\begin{figure}[htp]
	\centering
 % , height=5cm
	\includegraphics[width=8cm]{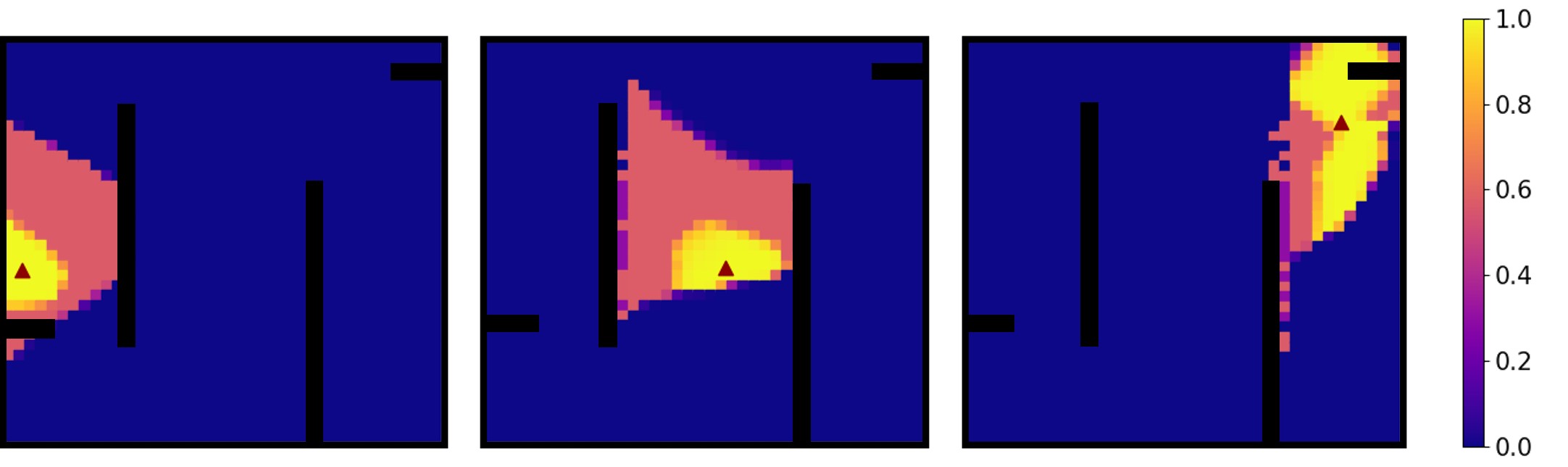}
        \caption{Heatmap of REM. The heatmap is based on the top view of \emph{UR-Pusher-2}. The red triangle represents the current state and different colors represent the reachability score between the two states. Warmer colors represent higher reachability scores. As is shown, REM can infer that the state behind the obstacle is difficult to reach within a small number of steps by the current policy, although the states are close in physical space.}
        
	\label{REM-result}
	 \vspace{-1.em}
\end{figure}

To answer the fourth question, we visualize the heatmap of REM, which shows the reachability score between two observations. Some examples are shown in Fig. \ref{REM-result}. The heatmap is based on the top view of \emph{UR-Pusher-2}. The red triangle represents the object position in the current state. Warmer colors represent higher reachability scores of the two states.
% Note that the input of REM is the disentangled representation of the robot and the object. 
To visualize the REM performance in the 2D plane, we keep the robotic end effector close to the object and uniformly sample the object position on the table. 

As is shown in Fig. \ref{REM-result}, the reachability of temporally close pairs is high, while that of temporally far pairs is low. In particular, when two states are separated by obstacles, even though they are close in physical space and representation space, REM can infer that the current GC policy cannot reach them within a small number of steps, and tend to predict these reachability scores to be zero.

\subsection{Ablation Study of Each Component in REPlan}\label{ablation_each}

\begin{table}[htp]\footnotesize
	\caption{Ablation Experiments for each component}
	 \vspace{-0.8cm}
	\label{table1}
	\begin{center}
		\linespread{1.2}\selectfont
		\begin{tabular}{l|c|c|c}
			
			\Xhline{1.5pt}
			
			\multirow{2}{*}{Method} & \multicolumn{3}{c}{Success rate(\%)}  \\
			\cline{2-4}
			& \emph{UR-Pusher-1} & \emph{UR-Pusher-2} & \emph{UR-Pick-Place} \\
			
			\Xhline{1.1pt}
			
			REPlan w/o REM\&CM & 0.0 & 0.0 & 0.0 \\
                \hline
                REPlan w/o CM & 56.7 & 36.7 & 30.0 \\
                \hline
                REPlan w/o DRM & 66.7 & 60.0& 53.3 \\
                \hline
                REPlan & 90.0 & 93.3 & 83.3 \\
			\hline
			\Xhline{1.5pt}
		\end{tabular}
	\end{center}
	\vspace{-1.6em}
\end{table}

To answer the fifth question, we perform several ablation experiments to analyze the role of different components in REPlan, which is shown in Table. \ref{table1}. The components include DRM, REM and Curiosity Module(CM). REPlan w/o REM\&CM replaces the subgoal metric with TDMs, which is similar to LEAP. REPlan w/o DRM uses VAE as the encoder. We sample 30 goals from the same distribution and test each model 30 times with 5 seeds. In one episode, when the Euclidean distance between the object and the goal in the plane is smaller than 0.03 meters, we define it as a success. 

As is shown, REPlan w/o REM\&CM cannot complete the tasks, since the model cannot explore high-quality samples in complex environments, leading to inaccurate distance measure and poorly planned subgoals. REPlan w/o CM can achieve some goals successfully, which shows that REM can help the model to plan better subgoals to guide the agent toward the final goal. However, the sample efficiency of the algorithm decreases in certain training epochs. We note that the success rate of REPlan w/o CM dropped faster in the more complex \emph{UR-Pusher-2} and \emph{UR-Pick-Place}. It shows that CM can further help agents to explore more high-quality samples in complex environments. 
REPlan w/o DRM makes certain progress in 2 tasks.
With all components, REPlan can reach the distant goals with a success rate of 90.0\% in \emph{UR-Pusher-1}, 93.3\% in \emph{UR-Pusher-2} and 83.3\% in \emph{UR-Pick-Place}. The further improvement of the success rate shows the compact disentangled representation extracted by DRM can significantly improve the performance of the algorithm.

\subsection{Real-world Experiments}

Our REPlan can effectively transfer control policies from simulation to the real world by separating scene representation and control. We transfer the GC policies learned in \emph{UR-Pusher-1} to the real-world task, named \emph{Real-Pusher}. We train DRM of REPlan with 500 real images, including different robot poses and object positions. The qualitative results are given in Fig. \ref{fenge-pic}. As is shown, our DRM can learn the disentangled representations of robots and objects from real observations.

\begin{figure}[t]
	\centering
	% , height=5cm
	\includegraphics[width=8.5cm]{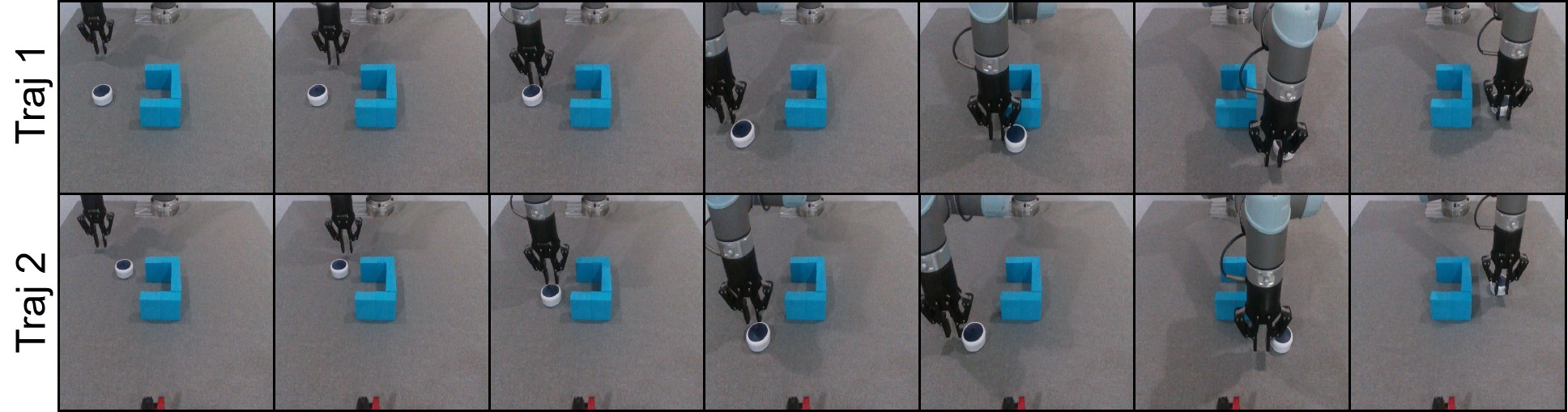}
	\caption{Some trajectories to successfully push the puck to the goal position in the real world.}
	\label{real-result}
	 \vspace{-1.6em}
\end{figure}

To align disentangled representations under different domains, we learn a simple linear mapping of robot representations with the same pose and object representations with the same position in simulated and real images. We test on UR5 30 times to push the puck to the goal position. Some successful trajectories are shown in Fig. \ref{real-result}. Despite some deviations in the representation mapping of different domains, our REPlan can reach the distant goals with a success rate of 76.67\% and exhibit transferability to the real world.

\section{Conclusion}
We propose a goal-conditioned algorithm combined with planning, denoted as disentanglement-based REachability Planning with Goal-Conditioned RL (REPlan).
REPlan decomposes temporally extended tasks into a sequence of subgoals to guide the GC policy to reach distant goals.
To explicitly extract task-relevant information, a self-supervised Disentangled Representation probabilistic Module (DRM) is proposed to disentangle the physical attributes of the entities (object positions and robot poses). On top of the representations, a REachability discrimination Module (REM) is designed to measure the reachability of subgoal sequences, which can further be optimized by a gradient-free planning algorithm. In addition, REM provides an intrinsic bonus to encourage exploration. We empirically demonstrate that REPlan can significantly outperform prior state-of-the-art methods in three challenging vision-based simulated tasks and exhibit transferability to the real world.

Our work suggests several directions for further research. While REPlan needs to collect offline datasets to learn the entity representations in the observations, how to simultaneously explore environments and learn representations remains challenging.
Self-supervised entity discovery from more cluttered real-world scenarios is another open issue. In addition, future work can encourage the exploration of GCRL by using self-generated subgoals or intrinsic bonuses to obtain high-quality samples for efficient planning and policy learning.

%\section*{Acknowledgments}
%This work was supported in part by the National Natural Science Foundation of China under Grant 62073244, Shanghai Innovation Action Plan under Grant 20511100500 and Innovation Program of Shanghai Municipal Education Commission (202101070007E00098).

\bibliographystyle{IEEEtran}     
\bibliography{reference}

\vfill

\end{document}